\documentclass[10pt, a4paper]{article}
\usepackage{lrec2022} 
\usepackage{multibib}
\newcites{languageresource}{Language Resources}
\usepackage{graphicx}
\usepackage{tabularx}
\usepackage{soul}
\usepackage{titlesec}
\titleformat{\section}{\normalfont\large\bfseries\center}{\thesection.}{1em}{}
\titleformat{\subsection}{\normalfont\SmallTitleFont\bfseries\raggedright}{\thesubsection.}{1em}{}
\titleformat{\subsubsection}{\normalfont\normalsize\bfseries\raggedright}{\thesubsubsection.}{1em}{}
\renewcommand\thesection{\arabic{section}}
\renewcommand\thesubsection{\thesection.\arabic{subsection}}
\renewcommand\thesubsubsection{\thesubsection.\arabic{subsubsection}}
\usepackage{epstopdf}
\usepackage{hyperref}
\usepackage{xstring}
\usepackage{color}

\usepackage{textcomp}
\usepackage{float}
\usepackage[inline]{enumitem}
\usepackage{times}
\usepackage{latexsym}
\usepackage{multirow}
\usepackage{appendix}
\usepackage{xspace}
\usepackage{scrextend}
\usepackage[utf8]{inputenc}
\usepackage[T2A,T1]{fontenc}
\usepackage{microtype}
\usepackage[russian,arabic,greek,main=english]{babel}
\def\thename{ID++\xspace}

\title{Don't Forget Cheap Training Signals Before Building Unsupervised Bilingual Word Embeddings}

\name{Silvia Severini, Viktor Hangya, Masoud Jalili Sabet, Alexander Fraser, Hinrich Sch{\"u}tze} 

\address{Center for Information and Language Processing \\
	LMU Munich, Germany \\
	\{silvia, hangyav, masoud, fraser\}@cis.uni-muenchen.de\\}

\abstract{
Bilingual Word Embeddings (BWEs) are one of the cornerstones of cross-lingual transfer of NLP models.
They can be built using only monolingual corpora without supervision leading to numerous works focusing on unsupervised BWEs.
However, most of the current approaches to build unsupervised BWEs do not compare their results with methods based on easy-to-access cross-lingual signals.
In this paper, we argue that such signals should always be considered when developing unsupervised BWE methods.
The two approaches we find most effective are: 1) using identical words as seed lexicons (which unsupervised approaches incorrectly assume are not available for orthographically distinct language pairs) and 2) combining such lexicons with pairs extracted by matching romanized versions of words with an edit distance threshold.
We experiment on thirteen non-Latin languages (and English) and show that such cheap signals work well and that they outperform using more complex unsupervised methods on distant language pairs such as
Chinese, Japanese, Kannada, Tamil, and Thai.
In addition,
they are even competitive with the use of high-quality lexicons in supervised approaches.
Our results show that these training signals should not be neglected when building BWEs, even for distant languages.
 \\ \newline  }

\begin{document}

\maketitleabstract

\section{Introduction}

Bilingual Word Embeddings (BWEs) are useful for many cross-lingual tasks.
They can be built effectively even when only a small seed lexicon is available by mapping monolingual embeddings into a shared space.
This makes them particularly valuable for low-resource settings \cite{mikolov2013exploiting}.
In addition, unsupervised mapping approaches can build BWEs for some languages when no seed lexicon is available.
Various unsupervised methods have been proposed relying on the assumption that embedding spaces are isomorphic \cite{zhang2017adversarial,conneau2017word,artetxe2018robust,alvarez2018gromov,chen2018unsupervised,hoshen2018non,mohiuddin2019revisiting,Alaux2019,Dou2020,grave2019unsupervised,li2020simple}.
However, with one exception, none of them
compare their results with the widely available baseline of using identical words as seed lexicons.

It has been shown that identical word pairs of two languages can be used to build high quality BWEs \cite{smith2017offline,artetxe2017learning}.
However, they were only tested on language pairs with similar scripts.
The only exception is the work of \newcite{sogaard2018limitations}, who tested identical word pairs on English and Greek which use different alphabetical characters but the same numerals.
Regardless of these experiments, recent works still propose novel unsupervised approaches without considering such cheap training signals, at least as baseline systems \cite{mohiuddin2019revisiting,Alaux2019,Dou2020,grave2019unsupervised,li2020simple}.

In this paper however, we argue that such signals
should be used as a cheap and effective
baseline in the development of future unsupervised methods.
We define them cheap as they require widely available monolingual corpora only, e.g., Wikipedia dumps,
but no parallel data.
We study two approaches for extracting the initial seed lexicons to build BWEs without relying on expensive dictionaries.
(1) First, we leverage identical pairs as proposed by \newcite{smith2017offline,artetxe2017learning}.
Previous work assumed such pairs not to be available for language pairs with distinct scripts, hence the development of various unsupervised mapping approaches.
We show that, surprisingly, they do appear in large quantities in the monolingual corpora that we use, even for distinct-script pairs.
In contrast to \newcite{sogaard2018limitations}, we test identical word pairs on multiple language pairs with distinct scripts, including pairs using distinct numerals. 
In addition, we propose to (2) strengthen identical pairs by extending them with further easily accessible pairs based on romanization and edit distance, which exploits implicit links between languages in the form of approximate word transliteration pairs.

We focus on distant language pairs having distinct scripts for many of which unsupervised approaches have failed or had very poor performance so far. For instance, English to Chinese, Japanese, Kannada, Tamil, and Thai, which all obtain a score close to 0 on the Bilingual Dictionary Induction (BDI) task \cite{vulic2019we}.
We evaluate the two approaches on thirteen different non-Latin\footnote{We use \emph{(non-)Latin language} here as a short form for \emph{language standardly written in a (non-)Latin script}.} languages paired with English on BDI.
We compare our lexicons' performance with unsupervised mapping and the frequently used MUSE training lexicons 
\citelanguageresource{conneau2017word}
and show that our noisy word pairs make it possible to build BWEs for language pairs where unsupervised approaches failed before and give accuracy scores similar to high quality lexicons.

Our work calls into question -- at least for BDI -- the strong trend toward unsupervised approaches in recent literature, similarly to \newcite{vulic2019we},
given that cheap signals are (i) available and easy to exploit, (ii) sufficient to obtain performance similar to dictionaries based on parallel resources like MUSE and (iii) able to make up for the failure of unsupervised methods.
Finally, we analyze which lexicon properties impact performance and show that our lexicon outperform unsupervised methods also for non-English language pairs.
Our paper calls for the need to use easily accessible bilingual signals, such as identical and/or transliteration word pairs, as baselines when developing unsupervised BWE approaches.

\section{Unsupervised pair extraction}
We show that we can extract the seed lexicon needed for mapping systems without the need for labeled data, making up for the failure of unsupervised methods. 
First, we show that identical pairs do appear in corpora of distant languages and can be exploited.
Secondly, we propose a novel method to boost the identical pairs sets by extracting the initial seed lexicon without the need for any bilingual knowledge, starting from monolingual corpora, and using romanization and edit distance.

\subsection{Identical pair approach} \label{sec:identical}
When dealing with languages with different scripts, identical pairs would seem to be unlikely to occur, which is assumed by unsupervised mapping methods. \newcite{smith2017offline,artetxe2017learning} form dictionaries from identical strings which appear in both languages but limit their approach to similar languages sharing a common alphabet, such as European ones.
Similarly, 
\citelanguageresource{conneau2017word}
refrain from using such identical word pairs, assuming they are not available for distant languages.
An exception is the work of \newcite{sogaard2018limitations} which shows the presence of identical pairs between English and Greek, which share numerals only but not alphabetical characters.

However, we show that there are domains where these pairs are actually available in large quantity even for pairs with different scripts, including the use of different numerals; an example is Wikipedia: see the statistics of fastText Wikipedia embeddings 
\citelanguageresource{Bojanowski2017}
 in Table \ref{tab:idpairs}.
Most of these identical pairs are
punctuation marks and digits, non-transliterated named entities written in the Latin script, or English words (assumingly words of a title) which were not translated in the non-English languages. This is also true for language pairs not including English.
In this paper, we build BWEs based on these pairs and show that they are sufficient for good BDI results on distant language pairs with distinct scripts.

\begin{table}[]
\centering
\begin{tabular}{lr|lr|lr}
Lang  & ID     & Lang  & ID     & Lang  & ID     \\ \hline
ko-th$^*$ & 17K  &  ko-he$^*$ & 11K & he-th$^*$ & 15K \\ \hline
en-zh$^*$ & 62K & en-bn$^*$ & 31K &  en-ar$^*$ & 19K \\
en-th & 46K & en-hi$^*$ & 30K &  en-ru & 18K \\
en-ja & 43K & en-ta$^*$ & 23K &  en-he$^*$ & 17K \\
en-el & 35K & en-kn$^*$ & 21K &  en-ko$^*$ & 15K \\
en-fa$^*$ & 32K &       &    & &
\end{tabular}
\caption{
Number of identical pairs per language pair.
Language pairs using different digits as their official numerals, on top of different alphabetical characters, are indicated with $^*$.
}
\label{tab:idpairs}
\end{table}

\subsection{Romanization based augmentation (\thename)}\label{sec:id++}
Identical pairs are noisy and may appear in smaller quantities for certain corpora and language pairs (e.g., he-ko). 
We propose our romanization approach that builds the seed lexicon completely automatically and can augment the identical pairs set.
We exploit the concept of transliteration and orthographic similarity to find a cheap signal between languages (cf.\ \cite{riley2018orthographic,severini2020combining,severini2020lmu,severini2022towards}) and to take advantage of cognates \cite{chakravarthi2019comparison,laville2020taln}.
It consists of 3 steps at the end of which we add the identical pairs and run VecMap in a semi-supervised setting.

\paragraph{1. Source candidates}~
First, we generate a list of source language words, which are the candidates to be matched with a word on the target side.
We use the English Wikipedia
dumps\footnote{\hyperlink{https://dumps.wikimedia.org/}{https://dumps.wikimedia.org/} (01.04.2020)} as our monolingual corpus and apply
Flair \cite{akbik2018coling} to extract  Universal Part-of-Speech (UPOS) tags.
We collect all English
proper nouns (PROPN),
since names are often transliterated between languages.
The resulting English proper noun set consists of $\approx$800K words.

\paragraph{2. Target candidates}~
The language-specific target data is extracted from the vocabulary of the pre-trained Wikipedia fastText embeddings 
\citelanguageresource{Bojanowski2017}.
The sets are not pre-processed with a POS tagger assuming that such a tool is missing or perform poorly for low-resource languages.
Compared to the English proper noun set, the vocabularies are smaller: between 40K and 500K.
Then, we romanize the corpora to obtain equivalent words but with only Latin characters -- this supports the distance-based metrics in step (3).
We use Uroman 
\citelanguageresource{hermjakob2018out} 
for romanization.
Examples of romanization are \begin{otherlanguage*}{russian}kарл\end{otherlanguage*} (Russian)$\rightarrow$ carl and \begin{otherlanguage*}{greek}βαβυλών\end{otherlanguage*} (Greek) $\rightarrow$ babylon.
Uroman mainly covers 1-1 character correspondences and does not vocalize words for Arabic and Hebrew. In general, its romanization is not as accurate as the transliteration of a neural model. However, neural models need a training corpus of labeled pairs to work well, while Uroman only uses the character descriptions from the Unicode table,\footnote{\hyperlink{http://unicode.org/Public/UNIDATA/UnicodeData.txt}{http://unicode.org/Public/UNIDATA/UnicodeData.txt}}
manually created tables and some heuristics, supporting a large number of languages.

\paragraph{3. Candidate matching}~
To find the corresponding target word for an English noun,
the noun is compared with each (romanized) target word
based on their orthography.
The similarity of two words $w_1$ and $w_2$ is defined as $1
- \mbox{NL}(w_1, w_2)$, where NL is the Levenshtein distance \cite{levenshtein1966binary} divided by the length of the longer string.
We select a pair of words if the similarity is  $\geq 0.8$; this ensures a trade off between number of pairs and quality, based on manual investigation.
We use the Symmetric Delete algorithm to speed up computation, similarly to \cite{riley2018orthographic}.
It takes the lists of source and target words, and a constant $k$ and identifies all the source-target pairs that are identical after $k$ insertion or deletions.\footnote{We used minimum frequency and minimum length equal to 1, $k$ equals to 2.}
The final step is to look up, for each romanized target word, its original non-romanized form.

\begin{table}[t]
\centering
\resizebox{.48\textwidth}{!}{%
\begin{tabular}{l|rrrrr}
 & en-th    & en-ja    & en-kn  & en-ta & en-zh  \\ \hline \hline
\multicolumn{6}{c}{\textbf{Unsupervised}} \\ \hline
1. & 0.00 & 0.96 & 0.00 & 0.07            & 0.07 \\
2.   & 0.00 & 0.48 & 0.00 & 0.07            & 0.00 \\
3. & 0.00 & 0.00 & 0.00 & 0.00$^\diamond$\hspace{-1.6mm} & 0.00 \\ \hline \hline
\multicolumn{6}{c}{\textbf{Semi-supervised} \cite{artetxe2018robust}} \\ \hline

ID  & \underline{24.40} & 48.87 & 22.03 & 17.93             & \underline{ 37.00} \\
Rom. & 23.33 & 48.46  & 22.90   & 18.00             & 0.27               \\ 
\thename & 23.47 & \underline{49.14} & \underline{24.23} & 18.20 & 35.00 \\ \hline
MUSE 
& 24.33  & 48.73  & 23.78
& \underline{18.80} & 36.53 
\end{tabular}
}
\caption{acc@1 on BDI for unsupervised
(1:~\protect\newcite{artetxe2018robust}, 2:~\protect\newcite{grave2019unsupervised}, 3:~\protect\newcite{mohiuddin2019revisiting})
and semi-supervised
approaches for 5 languages for which unsupervised methods fail. The semi-supervised results are obtained using VecMap with three different initial lexicons: the identical pair set (ID), ID extended with romanization based pairs (\protect\thename) and the MUSE dictionary.
We show an ablation study as well, i.e., the romanized pairs only (Rom.).
Scores from \protect\newcite{mohiuddin2020lnmap} are marked with $^\diamond$.
}
\label{tab:unsup}
\end{table}

\section{Evaluation}

We evaluate our seed lexicons on BDI to show the quality of the BWEs obtained with them.
Recent papers \cite{marchisio2020does} show that there is a direct relationship between BDI accuracy and downstream BLEU for machine translation. Moreover, \newcite{sabet2020simalign} show that good-quality word embeddings directly reflect the performance also for extrinsic tasks like word alignment.
We use the VecMap tool to build BWEs since it supports both unsupervised, semi-supervised and supervised techniques \cite{artetxe2018robust}.
The semi-supervised approach is of particular interest to us since it performs well with small and noisy seed lexicons
by iteratively refining them.
VecMap iterates over two steps: embedding mapping and dictionary induction. The process starts from an initial dictionary that is iteratively augmented and refined by extracting probable word pairs from the BWEs built in the current iteration with BDI.
The method is repeated until the improvement on the average dot product for the induced dictionary stays above a given threshold.
We use pre-trained Wikipedia fastText embeddings 
\citelanguageresource{Bojanowski2017}
 as the input monolingual vectors, taking only the 200K most frequent words and using default parameters otherwise.
We compare the performance of VecMap using our lexicons with MUSE.
MUSE contains dictionaries for many languages and it was created using a Facebook internal translation tool 
\citelanguageresource{conneau2017word},
thus it can be considered as a higher quality cross-lingual resource based on parallel data.
Since Kannada is not supported by MUSE,  we use the dictionary provided by \newcite{anonym_kannada}.
We show $acc@1$ scores based on CSLS vector similarity calculated by the MUSE evaluation tool 
\citelanguageresource{conneau2017word}.\footnote{We follow \newcite{artetxe2018robust} work for comparison reasons and did not remove identical pairs from the test sets. However, overlaps between train romanized lexicons and test lexicons correspond to less than 1\%.
}  

Tables \ref{tab:unsup} and \ref{tab:otherlang} show accuracy for all language pairs considering English as the source; see Table~\ref{tab:mainresults_appendix} in Appendix B for the full table containing results in both directions.
Table \ref{tab:unsup} gives scores for language pairs for which unsupervised methods completely diverge (acc@1 $<1$). We report results for three unsupervised methods 
\cite{artetxe2018robust,mohiuddin2019revisiting,grave2019unsupervised}.
In contrast, using identical word pairs as lexicon (ID) or its extension with the romanizetion based pairs (\thename)
with VecMap leads to successful BWEs without any parallel data or manually created lexicons.
In addition, scores are even comparable to high-quality dictionaries like MUSE.
Looking at results for all language pairs in Table \ref{tab:unsup} and \ref{tab:otherlang}, our sets always obtain results comparable to MUSE (baseline dictionaries), with improvements for Arabic, Chinese, Russian and Greek.
In the unsupervised cases (Table \ref{tab:unsup}), both ID and \thename pair sets lead to an accuracy improvement of at least $17$ points.
\thename outperform ID for three of the five low-resource pairs and five out of eight high-resource pairs proving that the romanized pairs can indeed strengthen the identical pairs sets.
These results show that good quality BWEs can be built by relying on implicit cross-lingual signals
without expensive supervision or fragile unsupervised approaches.

\paragraph{MUSE test w/o proper nouns}~ The work of \newcite{kementchedjhieva2019lost} highlights that MUSE test sets contain a high number of proper nouns for German, Danish, Bulgarian, Arabic and Hindi.
Since our romanization augmentation is based on such names, we evaluate their performance on the subsets of MUSE test sets that don't contain proper nouns. 
We remove proper nouns using the list of names obtained in Section \ref{sec:id++} and evaluate the performance of all the approaches presented above.
The new sets contains 10\% less pairs on average.
Results are shown in Table \ref{tab:muse_experiment_appendix}, Appendix C.
The performance is similar to the one obtained on the original test sets, proving that our dictionaries and methods are not biased towards aligning word embeddings of proper nouns.

\begin{table}[t]
\centering
\begin{tabular}{l|r|rrr|r}
        & Unsup.    & ID      & Rom.   & \thename & MUSE\\ \hline
en-ar   & 36.30    & 40.27   & 39.33 & 40.20  & 39.87    \\
en-hi   & 40.20    & 40.47   & 39.60 & 40.20  & 40.33    \\
en-ru   & 44.80    & 49.13   & 48.87 & 49.53  & 48.80    \\
en-el   & 47.90    & 47.87   & 48.00 & 48.27  & 48.00   \\
en-fa   & 36.70    & 37.67   & 36.80 & 37.67  & 38.00   \\
en-he   & 44.60    & 44.47   & 44.53 & 44.67  & 45.00    \\
en-bn   & 18.20    & 19.87   & 19.80 & 20.13  & 21.60    \\
en-ko   & 19.80    & 27.92   & 28.40 & 28.81  & 28.94 
\end{tabular}
\caption{acc@1 on BDI for (best) unsupervised method and semi-supervised VecMap with different initial lexicons.
(full table in Appendix B, Table \ref{tab:mainresults_appendix}).}
\label{tab:otherlang}

\end{table}

\paragraph{Non-English centric evaluation}~
We analyze the performance of 
ID and \thename for language pairs that do not include English.
We use the test dictionaries from \newcite{vulic2019we} that are derived from PanLex \cite{baldwin2010panlex,kamholz2014panlex} by automatically translating each source language word into the target languages.
We run VecMap for all combinations of Korean, Hebrew, and Thai.
Romanized train lexicons are extracted by combining the languages through English (e.g., th-ko is obtained using en-th and en-ko), i.e., words are paired if their English translation is the same.
Table \ref{tab:noneng_pairs} shows results.
When Thai is involved, the unsupervised method fails as for English-Thai.
Both ID and \thename always outperform the respective unsupervised scores, and perform similar to higher-quality dictionaries.
Additionally, ID++ outperforms ID in 3 out of 6 cases.
These results demonstrate further the simplicity and high quality of our methods.

\paragraph{Romanized-only}~
We analyze the performance of romanized pair lexicons on their own. Line Rom. in Table \ref{tab:unsup} and \ref{tab:otherlang} shows that they obtain competitive results to the other two approaches, with improvements for Japanese, and perform similarly to MUSE dictionaries.
The only failure is for Chinese (en-zh) -- presumably because Chinese has a logographic script that does not represent phonemes directly, so romanization is less effective.
These results show that the romanized pairs on their own also represent strong signals that shouldn't be neglected. Moreover, they constitute a good alternative when identical pairs are not available is such quantities (e.g., corpora of religious domain, law field, or cultural-specific documents).

\paragraph{Impact of OOVs}~
We analyze the pairs used for the various sets (Appendix A, Table \ref{tab:usedpairs_appendix}).
We define OOVs as words for which there is no embedding available among the pre-trained Wikipedia fastText embeddings.
Our romanized sets contain a substantial number of OOVs. (The identical pair sets do not contain OOVs because words are extracted from the top 200K most frequent.)
The main reason for OOVs is that the selected English pair of a word is so rare that they do not have embeddings.
On the other hand, the high number of OOVs (and resulting reduction of usable pairs) has only a limited negative impact on the performance.

\begin{table}[!t]
\centering
\resizebox{.48\textwidth}{!}{
\begin{tabular}{l|r|rrr|r}
      & Unsup.    & ID     & Rom.   & \thename                & PanLex  \\ \hline
th-ko &  0.00       &  2.81  & \underline{3.37}  & 3.09   & 2.95   \\
th-he &  0.00       &  \underline{9.75}  & 0.00  & 8.86   & 10.13  \\ \hline
ko-th &  0.00       &  \underline{15.90} & 14.23 & 15.26  & 14.36  \\
ko-he &  14.62      &  15.68 & \underline{16.08} & 16.00  & 15.11  \\ \hline
he-th &  0.00       &  16.42 & 0.00  & \underline{16.54}  & 17.90  \\ 
he-ko &  14.30      &  \underline{15.39} & 15.15 & 15.09  & 16.06  \\ 
\end{tabular}
}
\caption{acc@1 on BDI for unsupervised and semi-supervised VecMap for all combinations of Korean, Hebrew, and Thai. PanLex are results obtained with training lexicons from \protect\newcite{vulic2019we} and semi-supervised VecMap.
\label{tab:noneng_pairs}}
\end{table}

\paragraph{Size of seed set and word frequency}~
We analyze the impact of the size of the initial romanized seed set and of word frequency.
Appendix A, Table \ref{tab:freqbin_appendix}, displays accuracy scores for MUSE and Romanized lexicons containing the $n \in\{25 ,1000\}$ least and most frequent word pairs.
Performance of VecMap applied to seed sets of size 25 is close to 0. The only exception is Russian, where the unsupervised approach already works well. Next, we investigate seed sets of size 1000 consisting of either the least frequent or the most frequent words. High-frequency seed sets give better results as expected. The effect is particularly strong for Tamil: the high-frequency set has performance close to the full set, whereas the low-frequency set is at $\leq$0.07.
The performance of MUSE seed sets of size 25 and romanized seed sets of size 1000 is similar, demonstrating the higher quality of MUSE. However, obtaining the romanized pairs is much cheaper.

\section{Conclusion}

We have analyzed two cheap resources for building BWEs which can alleviate the issues of 
unsupervised methods which fail on multiple language pairs.
We focused on a wide range of non-Latin languages paired with English.
(i) We exploited identical pairs that surprisingly appear in corpora of distinct scripts.
We showed that they can be used even when numerals are distinct in contrast to previous work.
(ii) We combined them with a simple method to extract the initial hypothesis set via romanization and edit distance.  With both approaches, we obtained results that are competitive with high-quality dictionaries.  Without using explicit cross-lingual signal, we outperformed previous unsupervised work for most languages and in particular for five language pairs for which previous unsupervised work failed. Our results question the strong trend towards unsupervised mapping approaches, and show that cheap cross-lingual signals should always be considered for building BWEs, even for distant languages.

\section*{Acknowledgments}
This work was funded by the European Research Council (grant \#740516, \#640550), the German
Federal Ministry of Education and Research (BMBF, grant \#01IS18036A), and the German Research Foundation (DFG; grant FR 2829/4-1).

\section{Bibliographical References}\label{reference}

\bibliographystyle{lrec2022-bib}
\bibliography{lrec2022-example,custom}

\section{Language Resource References}
\label{lr:ref}
\bibliographystylelanguageresource{lrec2022-bib}
\bibliographylanguageresource{languageresource}

\appendix

\section{Statistics}

In this section we show statistics on the language pairs analyzed and additional scores.
Table \ref{tab:usedpairs_appendix} presents the number of pairs for each set that are not OOVs in the fastText wiki embeddings \citelanguageresource{Bojanowski2017}.

\begin{table}[!ht]
	\centering
	\resizebox{.5\textwidth}{!}{%
    \begin{tabular}{l|r|rr@{\hspace{1mm}}c@{\hspace{1mm}}rr@{\hspace{1mm}}c@{\hspace{1mm}}r}
    & MUSE     & ID & \multicolumn{3}{c}{Romanized} & \multicolumn{3}{c}{\thename}\\ \hline
    en-th      & 6,799      & 46,653    & 10,721&/&53,804   &58779&/&101066\\
    en-ja      & 7,135      & 43,556    & 11,488&/&118,626  &54970&/&161848\\
    en-kn      & 1,552      & 21,090    & 12,888&/&59,207   &33843&/&80032\\
    en-ta      & 8,091      & 23,538    & 5,987&/&120,836  &29472&/&143990\\
    en-zh      & 8,728      & 62,289    & 6,360&/&41,829  &68597&/&103971\\
    en-ar      & 11,571     & 19,275    & 4,773&/&61,031  &24019&/&80115\\
    en-hi      & 8,704      & 30,502    & 16,180&/&73,553   &46557&/&103791\\
    en-ru      & 10,887     & 18,663    & 9,913&/&301,698   &28520&/&319688\\
    en-el      & 10,662     & 35,270    & 20,740&/&150,472   &55841&/&185244\\
    en-fa      & 8,869      & 32,866    & 10,226&/&85,210   &43019&/&117817\\
    en-he      & 9,634      & 17,012    & 4,005&/&40,258    &20977&/&57059\\
    en-bn      & 8,467      & 31,954    & 10,721&/&53,804   &42573&/&85532\\
    en-ko      & 7,999      & 15,518    & 9956&/&134156 &25344&/&149031
	\end{tabular}}
	\caption{Number of pairs used that are not OOVs in the fastText wiki embeddings compared to the full size of the sets. For MUSE full and identical pairs sets there are no OOVs.
	}
	\label{tab:usedpairs_appendix}
\end{table}

\begin{table*}[!ht]
\centering
\begin{tabular}{cc|rrrr|rrrr}
&  & \multicolumn{4}{c|}{MUSE}      & \multicolumn{4}{c}{Rom.} \\
&  & 25L   & 25H   & 1000L & 1000H & 25L   & 25H   & 1000L & 1000H \\ \hline
\multirow{2}{*}{en-ta} & $\rightarrow$ & 14.73 & 16.27 & 17.33 & 17.40 & 0.00  & 0.00  & 0.07  & 17.80 \\
                       & $\leftarrow$ & 16.48 & 18.35 & 22.44 & 23.44 & 0.00  & 0.00  & 0.00  & 21.57 \\
\multirow{2}{*}{en-fa} & $\rightarrow$ & 35.33 & 34.20 & 38.07 & 37.20 & 0.00  & 0.20  & 37.47 & 37.47 \\
                       & $\leftarrow$ & 41.73 & 42.60 & 44.14 & 44.21 & 0.07  & 0.13  & 42.40 & 43.40 \\
\multirow{2}{*}{en-zh} & $\rightarrow$ & 39.00 & 39.40 & 38.20 & 37.67 & 0.00  & 0.00  & 0.07  & 0.40  \\
                       & $\leftarrow$ & 32.93 & 34.47 & 34.33 & 34.40 & 0.00  & 0.00  & 0.07  & 0.60  \\
\multirow{2}{*}{en-ru} & $\rightarrow$ & 49.07 & 43.07 & 49.07 & 49.27 & 49.33 & 47.73 & 49.40 & 49.00 \\
                       & $\leftarrow$ & 65.93 & 60.60 & 65.93 & 66.13 & 65.80 & 64.47 & 65.60 & 66.40
\end{tabular}
\caption{acc@1 using $25$ or $1000$ pairs lower-frequency (L) and higher-frequency (H) sets for MUSE and our romanized only (Rom.) set.
}
\label{tab:freqbin_appendix}
\end{table*}

\section{Main results}
In Table \ref{tab:mainresults_appendix} there are the accuracy scores based on CSLS vector similarity calculated by the MUSE evaluation tool 
\citelanguageresource{conneau2017word}. We show the scores for thirteen language pairs in both directions. The first five pairs are the ones for which unsupervised methods fail. We show both unsupervised and semi-supervised VecMap performance with baselines dictionaries and our three sets.

\begin{table*}[ht]
	\centering
	\resizebox{.8\textwidth}{!}{%
	\begin{tabular}{ccc|rrr|r|rrr}
\multicolumn{3}{l|}{} & \multicolumn{4}{c|}{Baselines}    & \multicolumn{3}{c}{Our}    \\
\multicolumn{3}{l|}{} & \multicolumn{3}{c|}{Unsupervised} & \multicolumn{1}{c|}{Semi-sup.} & \multicolumn{3}{c}{Semi-sup.}    \\
 &     &      & 1     & 2      & 3   & MUSE     & ID & Rom.  & \thename \\ \hline
 
\multirow{2}{*}{1}        & \multirow{2}{*}{en-th} & $\rightarrow$    & 0.00           & 0.00    & 0.00    & \textbf{24.33}    & \textbf{24.40}  & 23.33  & 23.47  \\
   &    & $\leftarrow$  & 0.00    & 0.00    & 0.00    & \textbf{19.04}   & \textbf{ 19.92} & 17.96 & 19.85\\ \hline
   
\multirow{2}{*}{2}        & \multirow{2}{*}{en-ja} & $\rightarrow$                   & 0.96           & 0.48            & 0.00      & \textbf{48.73}      & 48.87  & 48.46 & \textbf{49.14}\\
   &     & $\leftarrow$    & 0.96   & 0.00  & 0.00   & \textbf{32.87}   & 33.22   & \textbf{34.80}  & 33.43\\ \hline
   
\multirow{2}{*}{3}        & \multirow{2}{*}{en-kn} & $\rightarrow$                   & 0.00           & 0.00            & 0.00     & \textbf{23.78$^*$\hspace{-1.6mm}} & 22.03    & 22.90  & \textbf{24.23} \\
 &    & $\leftarrow$   & 0.00           & 0.00            & 0.00     
 & \textbf{41.25$^*$\hspace{-1.6mm}}   & \textbf{43.04} & 42.50 & 41.79 \\ \hline
 
\multirow{2}{*}{4}        & \multirow{2}{*}{en-ta} & $\rightarrow$    & 0.07     & 0.07     & 0.00$^\diamond$\hspace{-1.6mm}   & \textbf{18.80}         & 17.93           & 18.00  & \textbf{18.20}\\
 &                        & $\leftarrow$                    & 0.07           & 0.00            & 0.00$^\diamond$\hspace{-1.6mm}  & \textbf{24.38}                              & \textbf{24.78}  & 23.51 & \textbf{24.78}\\ \hline
 
\multirow{2}{*}{5}        & \multirow{2}{*}{en-zh} & $\rightarrow$                   & 0.07           & 0.00            & 0.00     & \textbf{36.53}   & \textbf{37.00}  & 0.27   & 35.00\\
 &     & $\leftarrow$                    & 0.00           & 0.00            & 0.00    & \textbf{ 32.80}                             & \textbf{34.33} & 0.07 & 32.67\\ \hline \hline
 
\multirow{2}{*}{6}        & \multirow{2}{*}{en-ar} & $\rightarrow$                   & 33.60  & 7.67                   & 36.30$^\diamond$\hspace{-1.6mm} & \textbf{39.87}                         & \textbf{40.27}           & 39.33   & 40.20\\
 &     & $\leftarrow$   & 47.72 & 12.92   & 52.60$^\diamond$\hspace{-1.6mm}  & \textbf{54.48}                       & 54.42           & 54.42  & \textbf{54.62}\\ \hline
                          
\multirow{2}{*}{7}        & \multirow{2}{*}{en-hi} & $\rightarrow$    & 40.20 & 0.00  & 0.00$^\diamond$\hspace{-1.6mm}   & \textbf{40.33}  & \textbf{ 40.47}  & 39.60  & 40.20\\
  &  & $\leftarrow$  & \textbf{50.57} & 0.07   & 0.00$^\diamond$\hspace{-1.6mm}  & 50.50 & 49.77  & 49.90 & \textbf{50.10} \\ \hline
  
\multirow{2}{*}{8}        & \multirow{2}{*}{en-ru} & $\rightarrow$    & \textbf{48.80}  & 37.33 & 46.90$^\diamond$\hspace{-1.6mm}  & \textbf{48.80}   & 49.13    & 48.87  & \textbf{49.53} \\
 &    & $\leftarrow$     & \textbf{66.13} & 52.73    & 64.70$^\diamond$\hspace{-1.6mm}   & 65.67   & \textbf{66.13}           & 65.73 & 66.07 \\ \hline
                          
\multirow{2}{*}{9}        & \multirow{2}{*}{en-el} & $\rightarrow$   & 47.67  & 34.67   & 47.90$^\diamond$\hspace{-1.6mm} & \textbf{48.00}  & 47.87   & 48.00 & \textbf{48.27}\\
&  & $\leftarrow$  & 63.40 & 49.20   & \textbf{63.50$^\diamond$\hspace{-1.6mm}} & 63.33  & 63.27  & \textbf{64.40} & 63.47 \\ \hline
                          
\multirow{2}{*}{10}       & \multirow{2}{*}{en-fa} & $\rightarrow$   & 33.27  & 0.53  & 36.70$^\diamond$\hspace{-1.6mm}     & \textbf{38.00} & \textbf{37.67}   & 36.80  & \textbf{37.67}    \\
   &    & $\leftarrow$   & 39.99 & 0.40    & \textbf{44.50$^\diamond$\hspace{-1.6mm}}  & 43.47  & \textbf{43.67}     & 42.93  & 43.60 \\ \hline
                          
\multirow{2}{*}{11}       & \multirow{2}{*}{en-he} & $\rightarrow$           & 44.60 & 37.13   & 44.00$^\diamond$\hspace{-1.6mm}   & \textbf{45.00}  & 44.47   & 44.53 & \textbf{44.67} \\
 &   & $\leftarrow$ & 57.88 & 50.01 & 57.10$^\diamond$\hspace{-1.6mm} & \textbf{57.94} & \textbf{58.14} & 57.81  & 57.94\\ \hline
                          
\multirow{2}{*}{12}       & \multirow{2}{*}{en-bn} & $\rightarrow$  & 18.20   & 0.00   & 0.00$^\diamond$\hspace{-1.6mm}      & \textbf{21.60} & 19.87 & 19.80 & \textbf{20.13}\\
 &  & $\leftarrow$ & 22.19 & 0.00                   & 0.00$^\diamond$\hspace{-1.6mm}  & \textbf{28.46}  & 28.88   & 28.67  & \textbf{29.41}   \\ \hline
                          
\multirow{2}{*}{13}       & \multirow{2}{*}{en-ko} & $\rightarrow$           & 19.80  & 9.62  & 0.00  & \textbf{28.94}     & 27.92  & 28.40  & \textbf{28.81}\\
 &   & $\leftarrow$  & 24.37  & 13.83           & 0.00    & \textbf{34.09}  & 33.40           & 33.74 & \textbf{33.95}
\end{tabular}
	}	
		\caption{acc@1 for unsupervised methods
		(1: \protect\newcite{artetxe2018robust}, 2: \protect\newcite{grave2019unsupervised}, 3: \protect\newcite{mohiuddin2019revisiting})
		and semi-supervised VecMap with different initial lexicons: MUSE set, identical pairs dataset (ID), our romanized only sets (Rom.), and the union of identical and romanized pairs (\protect\thename). We show both forward ($\rightarrow$) and backward ($\leftarrow$) directions. In bold the best result for each pair of languages, for ``Baselines'' and ``Our''. \\
		Scores from \protect\newcite{mohiuddin2020lnmap} are marked with $^\diamond$.
		\\ $^*$Kannada is not supported by MUSE, so we use the dictionary provided by \protect\cite{anonym_kannada}. }
		\label{tab:mainresults_appendix}
	\end{table*}

\section{MUSE proper nouns removal}
Table \ref{tab:muse_experiment_appendix} shows results computed on the subsets of MUSE test sets that don't contain proper nouns. 
We remove proper nouns using the list of names obtained in Section \ref{sec:id++}
The new sets contains 10\% less pairs on average.

\begin{table*}[ht]
	\centering
	\resizebox{.7\textwidth}{!}{%
	\begin{tabular}{ccc|r|r|rrr}
\multicolumn{3}{l|}{} & \multicolumn{2}{c|}{Baselines}    & \multicolumn{3}{c}{Our}    \\
\multicolumn{3}{l|}{} & \multicolumn{1}{c|}{Unsup} & \multicolumn{1}{c|}{Semi-sup.} & \multicolumn{3}{c}{Semi-supervised}    \\
 &     &      &   & MUSE     & ID & Rom.  & \thename \\ \hline
 
\multirow{2}{*}{1}        & \multirow{2}{*}{en-th} & $\rightarrow$    & 0.00    & \textbf{27.21}   & \textbf{27.13} & 26.35  & 26.11  \\
&    & $\leftarrow$  & 0.00    & 18.93   & 19.83 & 18.25 & 19.83\\ \hline
   
\multirow{2}{*}{2}        & \multirow{2}{*}{en-ja} & $\rightarrow$   & 0.71  & \textbf{46.15} & 45.04  & 46.31 & \textbf{46.39} \\
&  & $\leftarrow$  & 0.56  & \textbf{39.14}  & 38.86   & \textbf{40.73} & 39.52\\ \hline
   
\multirow{2}{*}{3}        & \multirow{2}{*}{en-kn} & $\rightarrow$ & 0.00 &  \textbf{23.78$^*$\hspace{-1.6mm}} & 22.03    & 22.90  & \textbf{24.23} \\
 &    & $\leftarrow$   & 0.00  & \textbf{41.25$^*$\hspace{-1.6mm}}  & \textbf{43.04} & 42.50 & 41.79 \\ \hline
 
\multirow{2}{*}{4}        & \multirow{2}{*}{en-ta} & $\rightarrow$    & 0.08 & \textbf{20.12} & 19.35  & 18.97  & \textbf{19.43} \\
 &        & $\leftarrow$    & 0.08   & \textbf{24.60} & 24.60 & 23.71 & \textbf{25.00} \\ \hline
 
\multirow{2}{*}{5}        & \multirow{2}{*}{en-zh} & $\rightarrow$    & 0.07  &  \textbf{37.34} & \textbf{38.14}  & 0.07   & 35.74 \\
 &     & $\leftarrow$   & 0.00    & \textbf{32.48} & \textbf{34.83} & 0.00 & 32.48 \\ \hline \hline
 
\multirow{2}{*}{6}        & \multirow{2}{*}{en-ar} & $\rightarrow$   & 35.44  & \textbf{ 39.70} & \textbf{40.23 }& 39.24 & 40.15\\
 &     & $\leftarrow$   & 49.75  & \textbf{53.61} & 53.46 & 53.61 & \textbf{53.82}\\ \hline
                          
\multirow{2}{*}{7}        & \multirow{2}{*}{en-hi} & $\rightarrow$    & \textbf{42.49}  & 42.42 & \textbf{42.79} & 42.11  & 42.57 \\
  &  & $\leftarrow$  & 52.46  & \textbf{52.62} & 51.99 & 52.07 & \textbf{52.23} \\ \hline
  
\multirow{2}{*}{8} & \multirow{2}{*}{en-ru} & $\rightarrow$    & \textbf{45.64} & \textbf{45.64}  & 46.40   & 45.64 & \textbf{46.70} \\
 &   & $\leftarrow$  & \textbf{64.35}& 64.13   & 64.57  & 64.35 & \textbf{64.72} \\ \hline
                          
\multirow{2}{*}{9}   & \multirow{2}{*}{en-el} & $\rightarrow$   & 48.90 & \textbf{49.35}& 48.97  & 49.43 & \textbf{49.58} \\
&  & $\leftarrow$  & \textbf{63.87} & 63.80  & 63.87  & \textbf{64.56} & 63.72 \\ \hline
                          
\multirow{2}{*}{10}       & \multirow{2}{*}{en-fa} & $\rightarrow$   & 34.18  & \textbf{37.51} & 37.35 & 36.58 & \textbf{37.59}  \\
   &    & $\leftarrow$   & 41.78 & \textbf{43.59} & \textbf{44.06} & 43.35 & 43.82 \\ \hline
                          
\multirow{2}{*}{11}       & \multirow{2}{*}{en-he} & $\rightarrow$    & 42.22 & \textbf{42.60} & \textbf{42.29} & 42.14 & \textbf{42.29} \\
 &   & $\leftarrow$ & \textbf{55.92} & 55.70 & 56.00 & 55.62  & \textbf{56.08} \\ \hline
                          
\multirow{2}{*}{12}       & \multirow{2}{*}{en-bn} & $\rightarrow$  & 20.44 & \textbf{22.74} & 21.59 & 20.52 & \textbf{20.98} \\
 &  & $\leftarrow$ & 25.80 & \textbf{30.22} & 30.30 & 30.30 & \textbf{30.96 }\\ \hline
                          
\multirow{2}{*}{13}       & \multirow{2}{*}{en-ko} & $\rightarrow$     & 20.30 & \textbf{26.57} & 25.63  & 26.02 & \textbf{26.49} \\
 &   & $\leftarrow$  & 26.52 & \textbf{32.37} & \textbf{32.21} & 31.80 & 32.13
\end{tabular}
	}	
		\caption{acc@1 on MUSE test sets without proper nouns.
		Results are reported for unsupervised and semi-supervised Vecmap \protect\newcite{artetxe2018robust} with different initial lexicons: MUSE set, identical pairs dataset (ID), our romanized only sets (Rom.), and the union of identical and romanized pairs (\protect\thename). We show both forward ($\rightarrow$) and backward ($\leftarrow$) directions. 
		In bold the best result for each pair of languages, for ``Baselines'' and ``Our''. \\
		$^*$Kannada is not supported by MUSE, so we use the dictionary provided by \protect\cite{anonym_kannada}. 
		}
		\label{tab:muse_experiment_appendix}
	\end{table*}

\section{Reproducibility} \label{sec:repr}
We run our method on up to 48 cores of Intel(R) Xeon(R) CPU E7-8857 v2 with 1TB memory and a single GeForce GTX 1080 GPU with 8GB memory. 
The training of semi-suprised BWEs using VecMap took approximately 1 hour per language pair.
For VecMap, as well as for all others methods we analyzed, we used the latest code available in their git repositories with default parameters.
\thename is implemented in Python.

\end{document}